\documentclass[runningheads]{llncs}
\usepackage{xurl}
\usepackage{graphicx}
\usepackage{booktabs}
\usepackage{hyperref}
\usepackage{xcolor}
\usepackage{amsmath}
\usepackage{microtype}
\usepackage[numbers,sort&compress]{natbib}
\newcommand{\myDots}{\ifmmode\mathinner{\ldotp\kern-0.13em\ldotp\kern-0.13em\ldotp}\else.\kern-0.13em.\kern-0.13em.\fi}

\begin{document}
\title{FairDistillation: Mitigating Stereotyping in Language Models}
\author{Pieter Delobelle\inst{1} \and
Bettina Berendt\inst{1,2}
\institute{Department of Computer Science, KU Leuven, Belgium\\
Leuven.AI institute, Belgium \and
Faculty of Electrical Engineering and Computer Science, TU Berlin, Germany \\
Weizenbaum Institute, Germany\\
}}
\maketitle              %

\abstract{%
Large pre-trained language models are successfully being used in a variety of tasks, across many languages. With this ever-increasing usage, the risk of harmful side effects also rises, for example by reproducing and reinforcing stereotypes. However, detecting and mitigating these harms is difficult to do in general and becomes computationally expensive when tackling multiple languages or when considering different biases. To address this, we present {\sc FairDistillation}: a cross-lingual method based on knowledge distillation to construct smaller language models while controlling for specific biases. We found that our distillation method does not negatively affect the downstream performance on most tasks and successfully mitigates stereotyping and representational harms. We demonstrate that FairDistillation can create fairer language models at a considerably lower cost than alternative approaches.
}


\section{Introduction}
Pre-trained transformer-based Language Models (LMs), like BERT~\citep{devlin2019bert}, are not only pushing the state-of-the-art across many languages, they are also being deployed in various %
services, ranging from machine translation to internet search~\citep{vaswani2017, devlin2019bert, liu2019roberta}. 
However, these deployed language models have been shown to exhibit problematic behaviour.
For instance, BERT and other models (i) replicate gender stereotypes~\citep{bartl2020, delobelle2020robbert,tan2019a}, (ii) exhibit dubious racial correlations~\citep{tan2019a} and (iii) reproduce racial stereotypes~\citep{may2019}. 
These behaviours are all present in pre-trained models that are used in a wide range of applications, which are referred to as \emph{downstream tasks}.

Without precautions, downstream tasks could use such problematic behaviour to make biased predictions.
LMs are generally finetuned for such tasks, where allocation harms (i.e. allocating or withholding a resource) might occur~\citep{blodgett2020language}. 
These can originate from the fine-tuning dataset or the pre-trained model or a combination of both.
We focus on the pre-training, where representation harms (i.e. encoding stereotypes) can occur in the pre-trained LMs~\citep{blodgett2020language, webster2020}.

Multiple methods have been proposed to reduce representational harms in language models~\citep{webster2020, bartl2020}, 
These methods are based on \emph{pre-processing} of the data, for example \emph{Counterfactual Data Augmentation} (CDA)~\citep{Lu2020} or \emph{Counterfactual Data Substitution} (CDS)~\citep{maudslay2019name}.
In both cases, gendered words in input sequences are replaced by a predefined counterfactual, e.g. ``\emph{He is a doctor}'' $\rightarrow$ ``\emph{She is a doctor}''.
CDA can significantly increase the training dataset, with longer training times as a consequence, so CDS-based methods replace input sequences instead.
Nevertheless, both techniques  
require retraining the model with an augmented dataset, instead of leveraging the efforts done to train the original model.

We propose a framework for mitigating representational harms based on knowledge distillation~\citep{hinton2015distilling}, which we demonstrate on gender stereotypes.
Our approach uses existing language models as a teacher, which provides a richer training signal and does not require retraining from scratch.
To prevent the transfer of learnt correlations to new LMs, our framework %
replaces CDA's augmentation strategy with probabilistic rules between tokens.
Since our approach can be performed at a fraction of the original training cost and also creates smaller models, it becomes more feasible to create domain-specific bias-controlled LMs.

In this paper, we start in \autoref{sec:related-work} with an overview of language models %
and fairness interventions (\autoref{ss:mitigate-lit}).
In \autoref{sec:fairdistillation}, we present our method to create debiased language models, which we call FairDistillation.
\autoref{sec:evaluation} describes the evaluation set-up and \autoref{sec:results} presents the results.
\autoref{sec:limitations} gives an overview of future work and ethical considerations and we conclude in \autoref{sec:conclusion}.

\section{Background}\label{sec:related-work}

BERT~\citep{devlin2019bert} is a language model that is trained in two phases: (i) self-supervised \emph{pre-training} with a Masked Language Modeling (MLM) objective and afterwards (ii) supervised \emph{finetuning} for downstream tasks.
The intuition behind the first learning task is that learning to reconstruct missing words in a sentence helps with capturing interesting semantics---and because this relies on co-occurrences it also captures stereotypes.
Formally, a token $x_m$ in the input sequence $x_1, \dots, x_N$ is replaced by a masked token ({\tt <mask>}) and the MLM objective is to predict the original token $x_m$ based on the context $\mathbf{x} = x_1, \dots, x_{m-1}, x_{m+1}, \dots, x_N$, following

\[
  \max_\theta \sum_{i=1}^N \mathbf{1}_{x_i = x_m} \log\left( P \left( x_i \mid \mathbf{x} ;\theta \right) \right)
\]

with $\mathbf{1}_{x_i = x_m}$ as an indicator function whether the token is correctly predicted.
This training setup results in a good estimator of the contextualized probability of a word $P\left( x_i \mid \mathbf{x} ;\theta \right)$. 
Aside from the MLM objective, the original BERT model also incorporated a Next Sentence Prediction (NSP) objective.
\citet{liu2019roberta} later concluded that the NSP objective did not improve training and removed it when constructing RoBERTa.
Because of this, we do not further consider this objective during distillation or evaluation.

After pre-training, the newly obtained model can be reused and finetuned for different classification and regression tasks, like sentiment analysis. 
Finetuning requires different datasets, that can also introduce biases that are referred to as \emph{extrinsic} biases~\citep{delobelle2021measuring}.
Mitigating extrinsic biases in downstream tasks is out of scope for this work.
Nevertheless, since LMs are used both for downstream tasks and for generating contextualized embeddings, mitigating intrinsic biases is still crucial.

\subsection{Mitigating intrinsic biases}\label{ss:mitigate-lit}
\citet{bolukbasi2016} presented two intrinsic debiasing methods based on removing the observed \emph{gender axis} in static word embeddings.
Mitigating problematic correlations is more challenging for LMs because of the contextualization that models like BERT incorporate.
This means that word representations from LMs cannot be considered in isolation, so mitigation strategies for word embeddings cannot be applied.

Models like BERT can only generate meaningful representations for a given sequence, so for this reason, mitigation strategies have mostly been based on Counterfactual Data Augmentation (CDA)~\citep{maudslay2019name, webster2020, bartl2020, Lu2020}.
This strategy augments the pre-training dataset with sequences where certain words, like pronouns or names, are swapped.

Unfortunately, this requires re-training the model from scratch, which can be extremely costly and with many negative side effects~\citep{bender2021dangers}.

One of few mitigation strategies that does not alter the training data was also presented by \citet{webster2020}, namely using dropout as a regularisation method against problematic correlations.
Regularisation as a means to mitigate problematic correlations thus seems a feasible option, but albeit effective, the method still requires retraining the model from scratch.
It should also be noted that these efforts are mostly focused on English.
Results of performing CDA on a German model were less successful, likely due to gender marking~\citep{bartl2020}.

All previous methods require retaining a language model.
\citet{lauscher-etal-2021-sustainable-modular} presents a unique approach, ADELE, that addresses this issue by using adapters~\citep{pmlr-v97-houlsby19a, pfeiffer-etal-2021-adapterfusion, pfeiffer-etal-2020-adapterhub}. 
These adapters are inserted after each attention layer and are the only trainable parameters, so the majority of parameters of a language model are shared over different tasks.
ADELE trains these adapters on a subset (1/3th) of the original BERT corpus\footnote{The original BERT corpus is a concatenation of Wikipedia and the Toronto Bookcorpus~\citep{devlin2019bert}} with the MLM objective and CDA to mitigate biases.
Although ADELE works very different from our distillation method, both methods aim to reduce the computational requirements and associated costs, by reusing existing models.

\subsection{Knowledge distillation}\label{ss:teacher}
Knowledge distillation is a method to transfer learnt knowledge from one model---originally proposed as an ensemble of models---to another, usually smaller model~\citep{hinton2015distilling,bucilua2006model}.
\citet{bucilua2006model} introduced this technique as model compression with an ensemble of models that are used to label a dataset.
This was later adapted for neural networks~\citep{hinton2015distilling}.
The teacher outputs a label probability distribution $z_i$ where some labels have higher probabilities, for example names or pronouns are more likely than verbs in the sentence ``{\tt <mask>} is a doctor.''.
To incorporate this information, a variation of the softmax function (\autoref{eq:softening}) can be used with a temperature $T$ to tune the importance of these labels.

\begin{equation}  
  p_i = \frac{\exp{\left(\frac{z_i}{T}\right)}}{\sum_j \exp{\left(\frac{z_j}{T}\right)}}.
  \label{eq:softening}
\end{equation}

\citet{sanh2019distilbert} focus on the distillation of the MLM task from pre-trained LMs.
Their models, DistilBERT and DistilRoBERTa, are trained on a linear combination of a distillation loss $\mathcal{L}_{ce}$ with the softening function from \autoref{eq:softening}, the original MLM loss $\mathcal{L}_{mlm}$, and additionally a cosine loss $\mathcal{L}_{cos}$ for the last hidden states, following

\[
  \mathcal{L} = \alpha_{ce} \mathcal{L}_{ce} + \alpha_{mlm} \mathcal{L}_{mlm} + \alpha_{cos} \mathcal{L}_{cos}.
\]

TinyBERT \citep{jiao2020tinybert} takes the same approach but also proposes a set of loss functions that perform distillation on (i) the embeddings layer, (ii) each of the transformer layers, and (iii) the prediction layer for specific tasks.
These different loss functions make TinyBERT perform slightly better than DistilBERT, but these functions require additional transformations to be learnt. 
In addition, if the student and teacher have a different number of layers, a mapping function is also required to transfer the knowledge between both.

\begin{figure*}[t]
    \centering
    \includegraphics[width=0.7\linewidth]{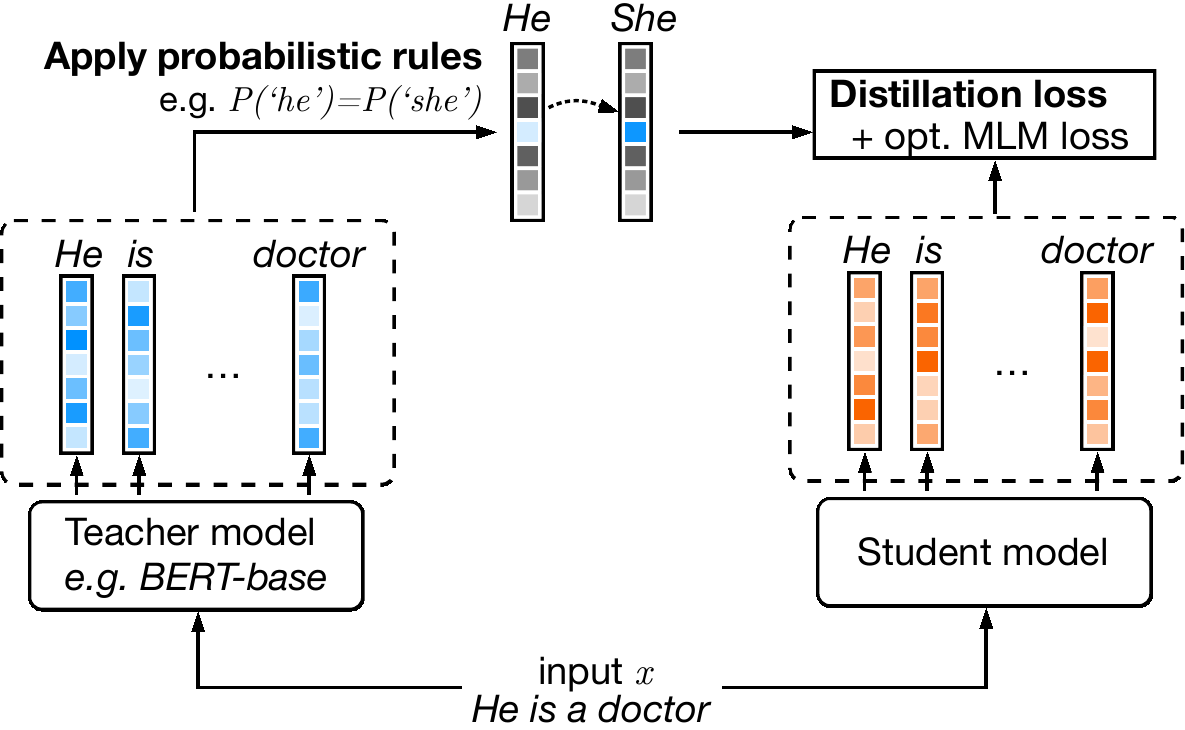}
    \caption{Overview of the training procedure with {FairDistillation} for a single input sequence in English.}
    \label{fig:architecture}
\end{figure*}

\section{FairDistillation}\label{sec:fairdistillation}
In this section, we introduce {FairDistillation}, a method to mitigate problematic correlations in pre-trained language models.
We first present the distillation architecture (\autoref{ss:architecture}) and afterwards, we will discuss the probabilistic rules that our method relies on (\autoref{ss:probabilistic-rules}).

\subsection{Architecture}\label{ss:architecture}
Our method trains a newly initialized model (the student) from an already trained model (the teacher).
Often, the teacher model has already been evaluated for biases, for example stereotypical gender norms for professions, which can lead to representational harms~\citep{blodgett2020language}.
In this example, models like BERT-base~\citep{devlin2019bert} predict that the input sentence ``\emph{{\tt <mask>} is a doctor.}'' should be filled with `\emph{He}' instead of `\emph{She}'.
The LM encoded that the token `\emph{He}' is more frequent in the training dataset, both in isolation and in combination with words like `\emph{doctor}'.

To prevent representational harms from being encoded in the final model, we apply a set of user-specified rules to the predictions of the original model.
By doing so, we can train a new model with these predictions.
Predictions of a teacher model provide a richer training signal and thus 
require less training time compared to CDA and CDS~\citep{sanh2019distilbert,hinton2015distilling,jiao2020tinybert}.
Moreover, we can simultaneously reduce the student's model size to improve both training and inference times, which boils down to knowledge distillation as is done for DistilBERT~\citep{sanh2019distilbert}.

\autoref{fig:architecture} illustrates our method, which consists of 5 steps.
First, an input sequence $x$ is passed to both the teacher and the student model, both with an MLM prediction head.
Second, the MLM predictions of the teacher model are passed to the rule engine.
Third, the predictions for certain tokens, like `\emph{He}', are modified based on the provided rules.
\autoref{fig:architecture} demonstrates how a rule where we assume equal probabilities $P(\text{`\emph{He}'}\mid x)=P(\text{`\emph{She}'}\mid x)$ alters the MLM prediction, which we discuss more in-depth later in this section.
Fourth, both MLM outputs, after applying possible rules to the teacher outputs, are used to calculate the distillation loss $\mathcal{L}_{ce}$ between the teacher and student outputs.
Finally, the MLM outputs of the student model can also be used to calculate an additional loss term $\mathcal{L}_{mlm}$ to train the student model in the same manner as the original model.

\paragraph{Student architecture.}
The student models use the same base architecture as the teacher models, but with 6 attention layers instead of the typical 12 layers, following \citet{sanh2019distilbert}.
The weights are initialized at random, which we prefer over smarter initialization strategies~\citep{sanh2019distilbert} to prevent an accidental transfer of problematic correlations.
We also reuse the teacher's tokenizer for the student, since these are already specifically constructed for the targeted language and no complex token translation is needed.

\paragraph{Applying probabilistic rules.}
The MLM head outputs a vector for each position in the input sequence, so for BERT-base this means at most 512 vectors. 
Each value in this vector represents the probability that a token fits in this position. Consequently, there will be 30,522 values for BERT-base-uncased.
We assume that some probabilities should be equal, like $P\left(\text{'He'} \mid \text{``{\tt<mask>} is a doctor''}\right) = P\left(\text{'She'} \mid \text{``{\tt<mask>} is a doctor''}\right)$, so our method can enforce these kind of equality rules.

During distillation, our method applies these equality rules to all the MLM outputs of the teacher.
For efficiency reasons, the tokens of interest are translated into a small lookup table at the start of the distillation loop so that applying each rule only requires a few lookup operations.
The corresponding values of the tokens are set to the mean of both values. Consequently, the outcome is also normalized and each prediction still sums up to 1.

Currently, our method only supports equalization between two or more tokens.
We did experiment with implementing these and more complex rules in ProbLog, a probabilistic logic programming language~\citep{deraedt2007problog}, but this proved to be unfeasible because of inference times that frequently exceeded 0.5 seconds per training example. 
Nevertheless, future work could focus on adding more complex rules that also depend for example on context or on part-of-speech tags to distinguish between adjectives (`His car' $\rightarrow$ `Her car') or pronouns (`\dots is his' $\rightarrow$ `\dots is hers'). %

\paragraph{Knowledge distillation.}
We follow the DistilBERT~\citep{sanh2019distilbert} distillation method, as discussed in \autoref{sec:related-work}.
FairDistillation applies a set of rules to affect the distillation loss, but %
the student not only learns from the distillation task, but also from the MLM task.
It is possible to concurrently train on this MLM objective for little additional cost.
Although this can be another source of problematic correlations, we opted to use this loss without correcting any associations. 
We reason that the contextual probability for a single input sequence can also be a useful signal. 

\subsection{Obtaining probabilistic rules}\label{ss:probabilistic-rules}
Until now, we used a running example of a probabilistic rule %
where the contextualized probability, as generated by the teacher LM, has to be equal for two tokens, namely `\emph{He}' and `\emph{She}'.
CDA achieves something similar by augmenting the dataset based on word mappings~\citep{Lu2020, feng2021survey}.
These mappings are very similar to our probabilistic rules; in fact, AugLy, a popular data augmentation framework~\citep{papakipos2022augly}, has the same mapping%
\footnote{\url{https://raw.githubusercontent.com/facebookresearch/AugLy/main/augly/assets/text/gendered_words_mapping.json}}
that we use for our running example in the context of gender bias.

Depending on which biases one wants to mitigate, different sets of rules are required.
We focus in this work on gender bias, so we rely on the same kind of rules as CDA. 
Simple rules to balance predictions highlight the robustness of our method and do not require lists of professions, which come with their own issues and biases~\citep{blodgett2021stereotyping}.
However, creating more fine-grained, domain-specific rules %
might improve our results. Such rules could aim at balancing, for example, profession titles or proper names.

\section{Experimental setup}\label{sec:evaluation}
We evaluate our method in two Indo-European languages: (i) English and (ii) Dutch, of which the results are discussed further in \autoref{sec:results}. 
Both languages have their own set of models, pre-training corpora and evaluation datasets, which we briefly cover in this section.
The evaluation of gender biases is also highly language-dependent and to illustrate generalization of our method beyond English, we also used a monolingual model for Dutch~\citep{delobelle2020robbert} with an architecture similar to RoBERTa~\citep{liu2019roberta}.
We opted for this language since it has some interesting, challenging characteristics, namely
it is one of only two languages with cross-serial dependencies that make it non-context free, with the other one being Swiss-German~\citep{bresnan1982cross}. 
It also has gendered suffixes for some, yet not all, nouns.
This affects such evaluations since these rely on implicit associations between nouns (e.g. for professions).
However, grammatical gender can also be an opportunity to evaluate how e.g. gendered professions align with the workforce \citep{bartl2020} or with equal opportunity policies.
We compare our method based on three popular metrics that we discuss in this section, \citet{delobelle2021measuring} provides a more comprehensive overview of intrinsic fairness measures.

\paragraph{SEAT.}
The Word Embedding Association Test (WEAT)~\citep{caliskan2017weat} measures associations between target words (`\emph{He}', `\emph{She}', \dots) and attribute words (`doctor', `nurse', \dots). 
Between the embeddings of each target and attribute word, a similarity measure like cosine similarity can be used to quantify the association between word pairs.
To add context, SEAT uses some `semantically bleached' template sentences~\citep{may2019}.

\paragraph{LPBS.}
\citet{kurita2019measuring} observe that using SEAT for the learned BERT embeddings fails to find many statistically significant biases, which is addressed in the presented \emph{log probability bias score} (LPBS). 
This score computes a probability $p_{tgt}$ for a target token~$t$ (e.g. `\emph{He}' or `\emph{She}') from the distribution of the masked position $X_m$ following
\[ 
  p_{tgt} = P\left(X_m = t \mid \mathbf{x}; \theta \right),
\]
for a template sentence, e.g. ``{\tt <mask>} is a doctor''.
Since the prior likelihood $P(X_m=t)$ can skew the results, the authors correct for this by calculating a template prior $p_{prior}$ by additionally masking the token(s) with a profession or another attribute $x_p$, following

\[ 
  p_{prior} = P\left(X_m = t \mid \mathbf{x} \backslash \{x_p\}; \theta \right).
\]
Both probabilities are combined in a measure of association $\log\frac{p_{tgt}}{p_{prior}}$ and the bias score is the difference between the association measures for two targets, like `\emph{He}' and `She`.
\citet{kurita2019measuring} applied their method to the original English BERT model~\citep{devlin2019bert} and found statistically significant differences for all categories of the WEAT templates.

\paragraph{DisCo.}
\citet{webster2020} also utilize templates to evaluate possible biases which their approach also mitigates (see \autoref{ss:mitigate-lit}).
As an intrinsic measure, the authors present \emph{discovery of correlations} (DisCo).
Compared to previously discussed metrics, this metric measures the difference in predictions for the attribute token $x_p$ when varying gendered tokens (i.e. \emph{`{\tt <P>} is a {\tt <mask>}'} for different pronouns or names instead of \emph{`{\tt <mask>} is a {\tt<P>}'} with different professions).

We experimented with the original DisCo metric, which performs statistical tests between predicted tokens, but we found that it didn't produce any statistically significant tokens. So, we simplified the metric to measure the differences in probabilities for the predicted tokens.
The resulting score of our DisCo implementation can therefore also be negative, while the original version has a lower bound of 0 as it counts the number of statistically significant fills. 

In the remainder of this section, we discuss our evaluations of English and Dutch in their respective subsections, where we define the used datasets, models and language-specific evaluation aspects. %

\subsection{English setup}
The first model we use as a teacher is the original uncased BERT model ({\tt BERT-base-uncased}) as released by \citet{devlin2019bert}, which is also the most-studied LM with regard to gender stereotypes.
This model was trained on the Toronto Bookcorpus and Wikipedia, but the Toronto Bookcorpus is no longer publicly available anymore and thus hinders reproduction.
For this reason, \citet{jiao2020tinybert} use only Wikipedia. %
We used a portion of the English section of the OSCAR corpus~\citep{ortiz-suarez-etal-2020-monolingual} to keep the training dataset size similar. More specifically, we used the first two shards of the unshuffled version.
We recognise that there is a mismatch between the domains of the Bookcorpus and OSCAR, but we believe this is acceptable to increase reproducibility.

As introduced in \autoref{ss:probabilistic-rules}, we use a set of gendered pronouns and define which ones should have the same probability. 
Since we use the uncased variant, we only need to define one set of rules, since `\emph{She}' and `\emph{she}' result in the same token.

The tokenization method used by BERT, WordPiece, splits words and adds a merge symbol (e.g. `word' + `{\tt\#\#}piece'), so no special care is required. 
For RoBERTa~\citep{liu2019roberta}, which uses Byte Pair Encoding (BPE), a word boundary symbol is used. 
Consequently, a word can have different tokens and representations depending if a space, punctuation mark, mask token or sequence start token are in front of the target token.
Since the Dutch model uses BPE, we will revisit this issue in \autoref{ss:dutch}.

\paragraph{Evaluation.}
To evaluate our method's performance trade-off, we finetune the obtained model on the Internet Movie Database (IMDB) sentiment analysis task~\citep{maas2011imdb}, which was also done by BERT~\citep{devlin2019bert} and DistilBERT~\citep{sanh2019distilbert}.
The dataset contains 25k training examples, from which we used 5k as a separate validation set, and another 25k test sequences. 
This is a high-level task where no gendered correlations should be used for predictions. 
Predicting entailment is a high-level task covered multiple times in GLUE~\citep{wang2019glue}, on which we also evaluated our method with the pre-trained model.
For a description of this benchmark and all datasets, we refer to \citet{wang2019glue}.

\paragraph{Bias evaluation.}
To evaluate possible problematic correlations with regard to gender stereotypes, we compute DisCo and LPBS, which we introduce at the beginning of this section.
We use the Employee Salary dataset\footnote{\url{https://github.com/keitakurita/contextual_embedding_bias_measure/blob/master/notebooks/data/employeesalaries2017.csv}}~\citep{kurita2019measuring}.
Following \citet{kurita2019measuring}, we filter on the top 1000 highest-earning instances as a proxy for prestigious jobs and test this for the same two templates (`{\tt <mask>} is a {\tt <P>}' and `{\tt <mask>} can do {\tt <P>}').
However, we additionally filter digits from the job titles and remove duplicate titles, to not skew the results towards more popular professions.

\subsection{Dutch setup}\label{ss:dutch}
We use a Dutch RoBERTa-based model called RobBERT~\citep{delobelle2020robbert} as a teacher, more specifically {\tt robbert-v2-dutch-base}.
This model was pre-trained on the Dutch section of the shuffled version of OSCAR\footnote{\url{https://oscar-corpus.com}\label{fn:oscar}}.
Similar to the distilled version of RobBERT~\citep{delobelle2022robbertje}, we select a 1GB portion of the OSCAR corpus (using {\tt head}, 2.5\%) to illustrate the ability to perform successful knowledge distillation with only a small fraction of the data required in comparison to the pre-trained model.

To create our model, we used a defined a set of rules based on the gendered pronouns `Hij' and `Zij' (`\emph{He}' and `\emph{She}').
The tokens corresponding to these pronouns were grouped based on capitalization and included spaces, since the BPE tokenizer includes a word boundary character at the beginning of some tokens.
Our method then used these rules to equalise the distributions predicted by the teacher during distillation, which we performed for 3 epochs.
This took approximately 40 hours per epoch on a Nvidia 1080 Ti and a traditionally distilled model required the same time, indicating our method has very limited effect on training time.

\paragraph{Evaluation.}
We compare the model created with FairDistillation to RobBERT and RobBERTje~\citep{delobelle2020robbert, delobelle2022robbertje} on the same set of benchmark tasks: (i) sentiment analysis on book reviews (DBRD)~\citep{vanderburghMerits2019dbrd}, (ii) NER, (iii) POS tagging, and (iv) natural language inference with SICK-NL~\citep{wijnholds2021sicknl}.%
These tasks are fairly high-level sequence-labelling tasks that can exhibit allocational harms, such as the predictive difference for sentiment analysis that was illustrated by \citet{delobelle2020robbert}.

\paragraph{Bias evaluation.}
We also evaluate numerically using the LPBS and DisCo metrics, but the RobBERT LM has also been evaluated by the authors on gender stereotyping using a different technique. %
This evaluation technique is based on a set of templates and a translated set of professions\footnote{\url{https://people.cs.kuleuven.be/~pieter.delobelle/data.html}} from \citet{bolukbasi2016}.
These professions have a perceived gender (e.g. `actress' is a female profession and `surveyor' is neutral), which can be correlated with the predictions by the model.
The authors rank the tokens based on the predicted probability instead of using this probability directly.
Interestingly, a correlation was not considered problematic, but male pronouns were predicted higher on average, even for by definition female professions (e.g. `nun').
To compare these results, we recreate the same plot and report the Mean Ranking Difference (MRD).
We focus on the gendered pronouns `\emph{zij}' (`\emph{she}') and `\emph{hij}' (`\emph{he}') for our evaluation.

\section{Results}\label{sec:results}

In this section, we present the results of the experiments (\autoref{sec:evaluation}).
We discuss English (\autoref{ss:en-results}) and Dutch (\autoref{ss:nl-results}) results separately.
We also performed experiments on French using the CamemBERT model~\citep{martin2020camembert}, but we chose to ommit those results due to our limited understanding of the language, which we address further in \autoref{sec:limitations}.

To eliminate any possible effect from hyperparameter assignments on the results, we ran each finetuning training 10 times with random hyperparameter assignments.
We varied the (i) learning rate, (ii) weight decay, and (iii) the number of gradient accumulation steps to effectively scale the batch size while still fully utilizing the GPU.
The full set of hyperparameters is listed in \autoref{tab:hp-space} in the supplementary materials.
For the Dutch benchmarks and for the English IMDB, we select the best-performing model based on the \emph{validation} set and present the results on the held-out \emph{test} set. 
The results from the GLUE benchmark are from the dev set, which were also the results reported by \citep{sanh2019distilbert}.

Unless indicated otherwise, all training runs are done on a single Nvidia 1080 Ti with 11 GB VRAM. 
All models are also comparably sized, with 66M trainable parameters each. This is 50\% of the model size of the teachers.

\begin{table*}[t]
\centering
\caption{English results on IMDB (sentiment analysis),  GLUE~\citep{wang2019glue}, and two bias measures. Following \citet{devlin2019bert}, we report $F_1$ scores for QQP and MRPC, Spearman correlations for STS-B, and accuracy for all other tasks. Results reported by \citet{devlin2019bert} on the GLUE dashboard are indicated with an obelisk ($^\dagger$), while the results from \citep{sanh2019distilbert} are also on the GLUE dev set, indicated with an asterisk ($^*$).
For LPBS, positive values represent more stereotypical associations, and for DisCo, lower values are more favorable.} 
\label{tab:results-en}
\resizebox{\textwidth}{!}{%
\begin{tabular}{@{}llllllllllll@{}}
\toprule
         & \textbf{}                           &                & \multicolumn{7}{c}{\textsc {\textbf{GLUE}}}    & \multicolumn{2}{c}{\textsc {\textbf{Bias}}}      \\ \cmidrule(l){3-10} \cmidrule(l){11-12} 
\textbf{Model}           & \textbf{IMDB}  & \multicolumn{1}{c}{\textbf{MNLI}} & \multicolumn{1}{c}{\textbf{QQP}} & \multicolumn{1}{c}{\textbf{QNLI}} & \multicolumn{1}{c}{\textbf{SST-2}} & \textbf{CoLA} & \textbf{STS-B} & \textbf{MRPC} & \textbf{RTE} & \textbf{LPBS} & \textbf{DisCo} \\ \midrule
BERT \citep{devlin2019bert}                    & $93.5$         & 84.0$^\dagger$                         & 71.2$^\dagger$                       & 90.5$^\dagger$                                   &  93.5$^\dagger$               &  52.1$^\dagger$             &   85.8$^\dagger$             &      88.9$^\dagger$         &  66.4$^\dagger$      & 1.16 & -0.48 \\
DistilBERT \citep{sanh2019distilbert}                 & $92.82$        & 82.2$^*$                                         & 88.5$^*$                                  &  89.2$^*$                                 &  91.3$^*$                                  & 51.3$^*$              & 86.9$^*$                & 87.5$^*$              & 59.9$^*$    &  -0.27 & -0.55       \\
FairDistillation            & $85.5 \pm 0.4$ &  80.1                                        & 82.1                                 &     86.6                              & 90.6               & 38.5              & 84.0                &  85.1             & 59.6        & \textbf{-0.16} & \textbf{0.25}    \\ \bottomrule
\end{tabular}%
}
\end{table*}

\subsection{English results}\label{ss:en-results}
We observe that problematic correlations are reduced on all three metrics, as is shown in \autoref{tab:results-en}.
One interesting observation---which also holds for Dutch---is that distillation in itself is already successful in mitigating these correlations.
This might be related to regularization as a method to control correlations~\citep{webster2020}, but we leave this for a future study.

On the IMDB task, our model suffers a 10\% accuracy drop, which is significant. 
However, as noted in \autoref{sec:evaluation}, we used a smaller training set for finetuning than BERT and DistilBERT, because we created a separate validation set from the original training set.

For GLUE, the results are in line with distilBERT. We do observe some diminished scores, notably CoLA, but the overall trade-off is limited.

Unlike the other models, we performed our FairDistillation method on 4 Nvidia V100's for 3 epochs, which took 70h per epoch. 
Finetuning was done on an Nvidia 1080 Ti for 4 epochs for IMDB, which took approximately 1h per run and was replicated 10 times.
For GLUE, we report the dev results and did not do any hyperparameter search.
We used the same hyperparameters as distilBERT~\citep{sanh2019distilbert}, who also report the development set results.

\subsection{Dutch results}\label{ss:nl-results}
Both the distilled RobBERT model and the model obtained with {\sc FairDistillation} perform only slightly worse (within 97.5\% of the original model) for both downstream tasks (see \autoref{tab:results}).
Both models have only half the parameters compared to the original RobBERT model and are thus faster to train and deploy, making this a decent trade-off between model size and predictive performance.
With no significant differences in performance between the distilled model and our {\sc FairDistillation} model, this highlights the potential of our method. 

\begin{table*}[t]
\centering
\caption{Dutch results on several benchmarks, namely Dutch Book Reviews (DBRD, sentiment analysis), named entity recognition (NER), part-of-speech (POS), tagging, and language inference (SICK-NL). We report bias as measured with LPBS and DisCo and additionally the mean ranking difference (MRD), which measures the preference of a language model to fill in male tokens (negative score) or female tokens (positive score). %
Benchmarks are reported with accuracy with 95\% CI, except for the NER task, where we report the $F_1$ score. Results indicated with $^\dagger$ were reported by \citet{delobelle2020robbert}.
For MRD, smaller ranking differences are more favorable, for LPBS, positive values represent more stereotypical associations, and for DisCo, lower values are more favorable.}
\label{tab:results}
\resizebox{\textwidth}{!}{%
\begin{tabular}{@{}lrlllllll@{}}
\toprule
              &                                                                        & \multicolumn{4}{c}{\textsc {\textbf{Benchmark scores}}} & \multicolumn{3}{c}{\textsc {\textbf{Bias}}}                                                                                      \\ \cmidrule(l){3-6} \cmidrule(l){7-9} 
\textbf{Model}&  \multicolumn{1}{c}{\textbf{Params}} & \multicolumn{1}{c}{\textbf{DBRD}}  & \multicolumn{1}{c}{\textbf{NER}} & \multicolumn{1}{c}{\textbf{POS}}  & \multicolumn{1}{c}{\textbf{SICK-NL}}  & \textbf{MRD}  & \textbf{LPBS} & \textbf{DisCo}\\ \midrule
RobBERT~\citep{delobelle2020robbert}       & 116 M           & $ 94.4 \pm 1.0^\dagger$                               & $89.1^\dagger$             & $96.4\pm 0.4^\dagger$ & $84.2 \pm 1.0$     & -7.47                                                             & 1.13                         & -0.29             \\ 
RobBERTje~\citep{delobelle2022robbertje}     & 74 M            & $92.5 \pm 1.1$         & $82.7$             &$\mathbf{95.6} \pm 0.4$ & $\mathbf{83.4} \pm 1.0$                   & -6.66                                                             & \textbf{-0.45}                         & -0.41            \\
FairDistillation            & 74 M            & $92.1 \pm 1.1$    &  $82.7$           & $95.4 \pm 0.4$      &  $82.4 \pm 1.1$      &    \textbf{-3.98}                                                    & 1.14                         & \textbf{-0.08}     \\
  \bottomrule

\end{tabular}%
}
\end{table*}

\begin{figure}[t]
    \centering
    \includegraphics[width=0.55\linewidth]{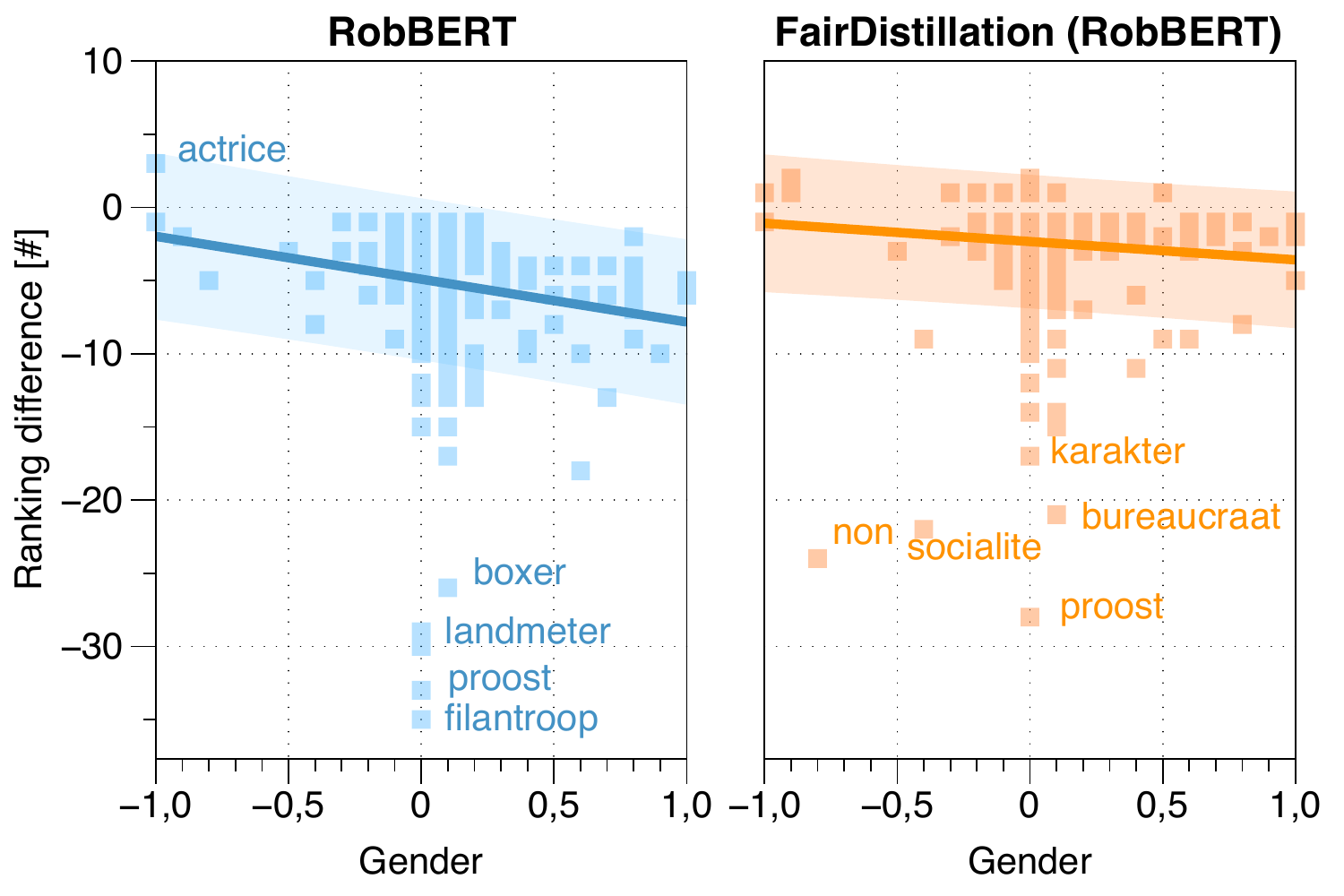}
    \caption{Differences in predictions for the Dutch template `{\tt <mask>} \emph{is een} {\tt <P>}' for our and the original RobBERT model. The `gender' axis ranges from words associated with female professions (left) to words associated with male professions (right). A positive ranking difference indicates `\emph{She}' is predicted before `\emph{He}'.}
    \label{fig:fairness-robbert}
\end{figure}

With regards to the bias evaluation, we observe a reduction between the original model and ours (\autoref{tab:results}): correlations are significantly reduced as measured by DisCo and the mean ranking of female associated tokens improved by 3.5 tokens. 
The only exception is LPBS~\citep{kurita2019measuring}, which incorporates a correction based on the prior probability of a token.
Our method effectively corrects this prior, while still allowing the context to affect individual results with the MLM objective.
Since Dutch has gendered nouns for some professions, a correlation is not necessarily undesirable, but the prior is (e.g. assuming all physicians are men).
Further graphical analysis of the predicted rankings for the third person singular pronouns confirms this, as shown in \autoref{fig:fairness-robbert}.
These charts reveal that most professions are now less associated with the masculine pronoun.
When considering which pronoun is ranked higher (i.e. above or below $y=0$), this result is even more pronounced. 
RobBERT only predicted a feminine pronoun for a single profession \citep{delobelle2020robbert}, while with our method this increased to 15 professions.

\section{Limitations and ethical considerations}\label{sec:limitations}
Despite the promising results, there are several potential improvements possible to our methods, as well as some ethical considerations. 
First, we rely on facts that express probabilities for a single token at a time. 
For gender stereotyping, this is sufficient as the vocabulary usually contains the tokens of interest.
However, this is not the case for many other problematic correlations, especially those affecting minority groups.
Tokens that are interesting here, like names, are not in the tokenizer's vocabulary because this is created based on occurrence counts.
Addressing this limitation would require extending our method to support facts that span multiple tokens.

Second, the effects of our method on `low-level' grammatical tasks require further study in English, as we focused on GLUE and sentiment analysis. The used Dutch benchmarks do cover more tasks and seem to indicate favourable reductions.

Third, our work focuses on binary gender stereotypes.
This leaves out a wide range of people who do not identify as such and although our method supports equalization over more than two tokens, this might be challenging if the intended words span multiple tokens. %
This also poses a challenge for generalizing our method beyond gender bias, since this frequently involves words that are not a single token in BERT's vocabulary. 

Fourth, since none of the authors is a native speaker of a language like French, we only performed a limited, exploratory evaluation of our methods with CamemBERT~\citep{martin2020camembert}.
Our method appears successful, with improved scores of 0.04 (DisCo) and -0.85 (LPBS) compared to -1.15 (DisCo) and 1.99 (LPBS) for the original CamemBERT model~\citep{martin2020camembert}. 
The performance of the LM was also still high, with XNLI~\citep{conneau2018xnli} scores 75.6 compared to  82.5 for CamemBERT.
However, constructing correct probabilistic rules and evaluating them is tricky for non-native speakers. %
For example, the female variant of a profession can refer to a woman practising said profession, but also to the spouse of a man with this profession.
When discussing these results with native French speakers from Wallonia, Belgium and from northern France, we realised that we are not well-suited to address this.
We thus leave a more comprehensive evaluation across languages as future work. 

Finally, by presenting a method to remove correlations with gender stereotypes in pre-trained language models, we risk it being used as a `rubber stamp' to absolve model creators from their responsibilities.
Therefore, we urge creators to critically analyse LMs within the social context that these models will be deployed in, both with respect to both the pre-trained and the finetuned model.

\section{Conclusion}\label{sec:conclusion}
We introduced a method called FairDistillation that allows to use probabilistic rules during knowledge distillation. 
We showed that this can effectively mitigate gender stereotypes in language models.
Our method demonstrates that knowledge distillation of language models with probabilistic rules is a possible alternative to re-training in order to reduce representational harms.
Even though comes at a slight cost for some downstream tasks, but we find that the overall cost is limited and can mostly be attributed to the distillation process.

\section*{Acknowledgements}
Pieter Delobelle was supported by the Research Foundation - Flanders (FWO) under EOS No. 30992574 (VeriLearn) and received funding from the Flemish Government under the ``Onderzoeksprogramma Artificiële Intelligentie (AI) Vlaanderen'' programme.
Bettina Berendt received funding from the German Federal Ministry of
Education and Research (BMBF) – Nr. 16DII113.
Some resources and services used in this work were provided by the VSC (Flemish Supercomputer Center), funded by the Research Foundation - Flanders (FWO) and the Flemish Government.

\bibliography{biblio}
\newpage
\appendix
\section{Hyperparameters}
\begin{table}[h]
\centering
\caption{The hyperparameter space used for finetuning.}
\label{tab:hp-space}
\begin{tabular}{@{}ll@{}}
\toprule
\textbf{Hyperparameter}         & \textbf{Value}         \\ \midrule
adam\_epsilon                   & $10^{-8}$              \\
fp16                            & False                  \\
gradient\_accumulation\_steps   & $i \in \{1, 2, 3, 4\}$ \\
learning\_rate                  & $[10^{-6}, 10^{-4}]$   \\
max\_grad\_norm                 & 1.0                    \\
max\_steps                      & -1                     \\
num\_train\_epochs              & 3                      \\
per\_device\_eval\_batch\_size  & 4 (16 for XNLI)                      \\
per\_device\_train\_batch\_size & 4 (16 for XNLI)                      \\
max\_sequence\_length.          & 512 (128 for XNLI)    \\
seed                            & 1                      \\
warmup\_steps                   & 0                      \\
weight\_decay                   & $[0, 0.1]$             \\ \bottomrule
\end{tabular}
\end{table}

\end{document}